\documentclass[dvipsnames]{article} 
\usepackage{LLaDA-FastV_conference}

\usepackage{booktabs}
\usepackage{graphicx}
\usepackage{enumitem}
\usepackage{wrapfig}
\usepackage{algorithm}
\usepackage{algpseudocode}

\usepackage{microtype}
\usepackage{amsmath}
\usepackage{colortbl}
\usepackage[utf8]{inputenc}
\definecolor{lightgray}{rgb}{0.9,0.9,0.9}
\usepackage{caption}
\usepackage{subcaption}
\usepackage{setspace}
\usepackage{url}
\usepackage{multirow}
\usepackage{colortbl}
\usepackage{tabularx}
\usepackage{blindtext}
\usepackage{pgfplots}
\pgfplotsset{compat=1.18} 
\usepackage{tikz}
\usetikzlibrary{er,positioning,bayesnet}
\usepackage{makecell}
\usepackage{tipa}
\usepackage{siunitx}
\usepackage{nicefrac}
\usepackage{tocloft}
\usepackage{listings}
\usepackage[raster,skins]{tcolorbox} %
\usepackage{xltabular}
\usepackage{adjustbox}
\usepackage{xurl}
\usepackage{rotating}
\usepackage[normalem]{ulem}
\usepackage{xcolor} 
\usepackage{cuted}   
\usepackage{capt-of} 
\usepackage{float} 
\usepackage{setspace}
\usepackage{stfloats}

\useunder{\uline}{\ul}{}

\usepackage{amsmath,amsfonts,bm}

\def\eqref#1{equation~\ref{#1}}

\def\1{\bm{1}}

\DeclareMathAlphabet{\mathsfit}{\encodingdefault}{\sfdefault}{m}{sl}
\SetMathAlphabet{\mathsfit}{bold}{\encodingdefault}{\sfdefault}{bx}{n}

\newcommand*\justify{%
  \fontdimen2\font=0.4em
  \fontdimen3\font=0.2em
  \fontdimen4\font=0.1em
  \fontdimen7\font=0.1em
  \hyphenchar\font=`\-
}

\renewcommand{\texttt}[1]{%
  \begingroup
  \ttfamily
  \begingroup\lccode`~=`/\lowercase{\endgroup\def~}{/\discretionary{}{}{}}%
  \begingroup\lccode`~=`[\lowercase{\endgroup\def~}{[\discretionary{}{}{}}%
  \begingroup\lccode`~=`.\lowercase{\endgroup\def~}{.\discretionary{}{}{}}%
  \catcode`/=\active\catcode`[=\active\catcode`.=\active
  \justify\scantokens{#1\noexpand}%
  \endgroup
}

\title{Efficient Token Pruning for LLaDA-V}

\author{
\bf
Zhewen Wan\thanks{Corresponding author. Email: w\_g\_rem@sjtu.edu.cn},
Tianchen Song\thanks{Corresponding author. Email: songtianchen@lixiang.com},
Chen Lin,
Zhiyong Zhao,
Xianpeng Lang\\
Li Auto Inc.
}

\setcitestyle{numbers,square,comma}

\begin{document}

\maketitle

\begin{abstract}
Diffusion-based large multimodal models, such as LLaDA-V \citep{llada}, have demonstrated impressive capabilities in vision–language understanding and generation. However, their bidirectional attention mechanism and diffusion-style iterative denoising paradigm introduce significant computational overhead, as visual tokens are repeatedly processed across all layers and denoising steps. In this work, we conduct an in-depth attention analysis and reveal that, unlike autoregressive decoders, LLaDA-V aggregates cross-modal information predominantly in middle-to-late layers, leading to delayed semantic alignment. Motivated by this observation, we propose a structured token pruning strategy inspired by FastV \citep{fastv}, selectively removing a proportion of visual tokens at designated layers to reduce FLOPs while preserving critical semantic information. To the best of our knowledge, this is the first work to investigate structured token pruning in diffusion-based large multimodal models. Unlike FastV, which focuses on shallow-layer pruning, our method targets the middle-to-late layers of the first denoising step to align with the LLaDA-V’s delayed attention aggregation to maintain output quality, and the first-step pruning strategy reduces the computation across all subsequent steps. Our framework provides an empirical basis for efficient LLaDA-V inference and highlights the potential of vision-aware pruning in diffusion-based multimodal models. Across multiple benchmarks, our best configuration reduces computational cost by up to 65\% while preserving an average of 95\% task performance.
\end{abstract}

\begin{figure}[t] 
  \centering
  \includegraphics[width=\textwidth]{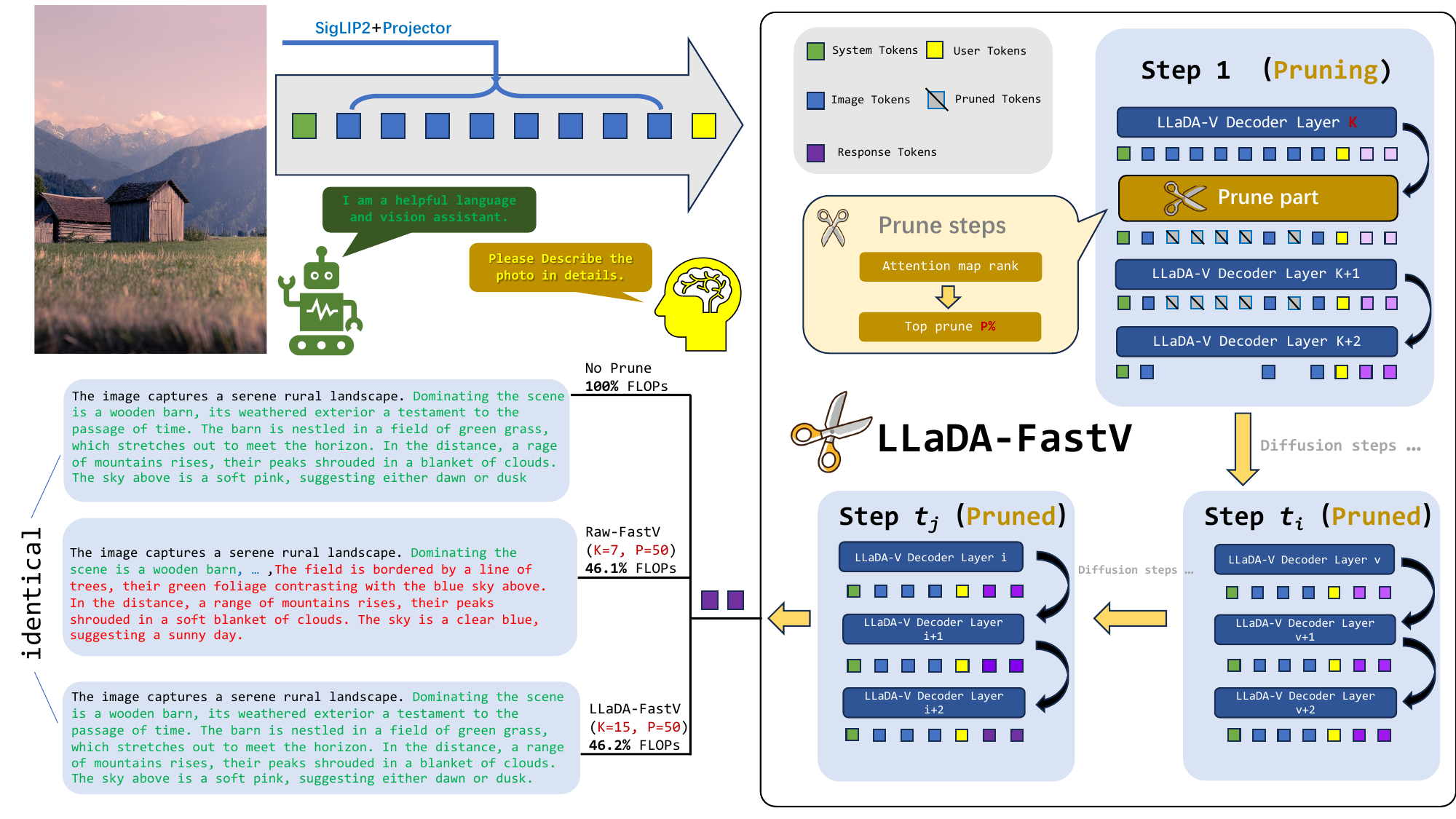}
  \caption{
    \textbf{Overview of LLaDA-FastV.} 
    We propose a token pruning framework tailored for iterative generative models. 
    \textbf{Right:} By applying FastV at Layer $K$ during the initial step (Step 1), redundant visual tokens (grey blocks) are permanently discarded. Subsequent generation steps (e.g., Step $t$) operate on the reduced sequence, significantly lowering computational costs. 
    \textbf{Left:} Performance comparison across different pruning layers. While aggressive early pruning ($K=7$) leads to hallucinations (highlighted in \textcolor{red}{red}), pruning at a deeper layer ($K=15$) maintains semantic fidelity comparable to the baseline (highlighted in \textcolor{green!60!black}{green}) while reducing FLOPs by $\sim 54\%$.
  }
  \label{fig:main_teaser}
\end{figure}

\section{Introduction}
\label{sec:intro}

In recent years, large-scale Vision-Language-Action (VLA) models have emerged as a promising paradigm for integrating perception, reasoning, and decision-making within a unified framework \citep{rt2, palme, llava}. Unlike conventional computer vision pipelines that rely on isolated modules for detection, tracking, and planning \citep{maskrcnn, yolox}, VLA models enable semantic abstraction and contextual reasoning across heterogeneous modalities. Among these, LLaDA-V represents a distinctive architecture that leverages a diffusion-based generative paradigm. By separating the visual encoder, language-based reasoning module, and action head into distinct yet coordinated sub-systems, LLaDA-V supports structured semantic reasoning over temporal and spatial observations, offering improved stability and extensibility for complex downstream tasks.

Autonomous driving presents a particularly compelling application scenario for such VLA models \citep{drivegpt4, uniad, vad}. Vehicles must continuously process multi-view visual inputs and reason about interactions among agents in dynamically evolving environments. Traditional perception–planning pipelines typically treat object detection and behavior reasoning as separate stages \citep{apollo, autoware}, limiting their ability to capture high-level semantics such as intention and risk awareness. By contrast, VLA models provide a unified representational space where visual perception and linguistic abstraction interact. For instance, LLaDA-V can ground latent concepts like ``a pedestrian shows crossing intent'' directly into the decision-making process, enhancing both interpretability and generalization to long-tail traffic scenarios.

Despite these advantages, deploying LLaDA-V in latency-critical environments remains a formidable challenge. The model typically operates on high-resolution visual inputs, resulting in a massive number of visual tokens. More critically, LLaDA-V employs an \textit{iterative generative mechanism} (e.g., re-masking or diffusion steps), where the heavy decoder layers are executed repeatedly for a single inference request. This introduces substantial redundancy, as the cross-modal attention mechanism repeatedly processes the same visual tokens across all layers and all time steps. In resource-constrained onboard scenarios, this leads to prohibitive inference latency and computational overhead, hindering practical deployment where real-time responsiveness is essential for safety \citep{edge_computing}.

These observations reveal a fundamental tension: while richer cross-modal representations improve semantic reasoning, they amplify the computational burden along the critical inference path. Recent studies in autoregressive VLMs have identified that visual tokens often contain significant redundancy \citep{vispruner, fastdrive, sparsevlm, divprune}. However, blindly applying existing pruning methods to LLaDA-V is suboptimal. Our in-depth analysis of LLaDA-V's attention maps reveals a unique phenomenon: unlike standard LLMs, LLaDA-V aggregates cross-modal information predominantly in the \textit{middle-to-late layers}. Aggressive pruning in early layers disrupts this feature aggregation, leading to severe hallucinations, while pruning in deeper layers preserves semantic fidelity, reducing the redundancy of calculation in the same time.

To address this challenge, we propose \textbf{LLaDA-FastV}, a structured token pruning framework tailored for iterative generative models. Unlike traditional FastV, which operates solely within the spatial dimension (i.e., layers) of autoregressive models, LLaDA-FastV introduces a \textit{spatiotemporal pruning} paradigm.

Our key insight is that iterative generation offers a unique temporal axis for optimization. By identifying and removing redundant visual tokens at a designated depth (spatial, e.g., Layer 15) during the initial generation step (temporal), we can enforce a \textit{Persistent Pruning} state. This ensures that the reduced token sequence is maintained across all subsequent denoising iterations. Consequently, the computational savings are effectively amortized over the entire generation process—yielding higher acceleration ratios as the number of denoising steps increases, while preserving model accuracy by pruning only after semantic features have sufficiently aggregated in the middle-to-late layers.

Our main contributions are summarized as follows:
\begin{itemize}
\item We conduct the first comprehensive attention analysis of the LLaDA-V architecture, revealing its delayed semantic aggregation pattern. This motivates our spatiotemporal design, where we identify the optimal layer (spatial) and time-step (temporal) to execute pruning without compromising semantic integrity.
\item We introduce a Persistent Pruning strategy that distinguishes our method from autoregressive counterparts. By performing pruning at the initial step and locking the token indices, we reduce FLOPs by approximately \textbf{54\%} (retaining only 46\% computation). Crucially, our method demonstrates superior efficiency scaling: the greater the number of denoising steps, the more significant the speedup relative to the baseline.
\item Extensive experiments on benchmarks such as AI2D, MMMU, and RealWorldQA demonstrate that our method maintains, and in some cases surpasses, the baseline performance, while significantly reducing latency for autonomous driving applications.
\end{itemize}
\section{Inefficient Attention in LLaDA-V}

Before introducing our acceleration and pruning strategy, we first conduct an internal attention visualization study on the original LLaDA-V model. We extract and visualize the attention weights from its cross-modal interaction stages and intermediate layers to examine how visual tokens participate in the reasoning process and how semantic contributions are distributed across different regions. By comparing attention mappings across layers and heads, we are able to observe tokens that consistently receive high attention and play a dominant role in semantic grounding, as well as tokens that are repeatedly involved in computation despite contributing limited task-relevant information.

\begin{figure}[htbp]
\centering
\includegraphics[width=0.95\linewidth]{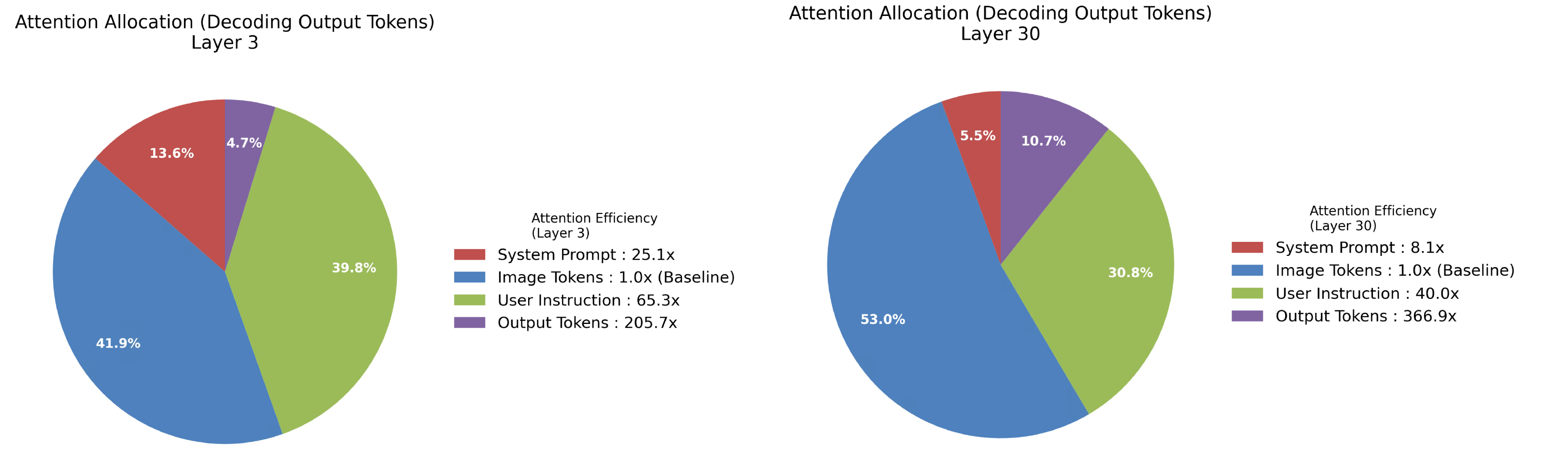}
\caption{Illustration of inefficient visual attention phenomena.}
\label{fig:attention-vis}
\end{figure}

This visualization analysis provides a qualitative perspective on the inherent attention behavior of LLaDA-V prior to acceleration. It reveals both the structural patterns of attention concentration and the presence of potential computational redundancy along the inference path. These observations form an empirical basis for our subsequent design of FastV-based token reduction and structured pruning, motivating the need to suppress semantically insensitive tokens while retaining those most associated with critical scene semantics and downstream reasoning objectives. The corresponding visualization results and detailed discussions are presented in Figure~\ref{fig:attention-vis}.

\begin{figure}[htbp]
\centering
\includegraphics[width=0.95\linewidth]{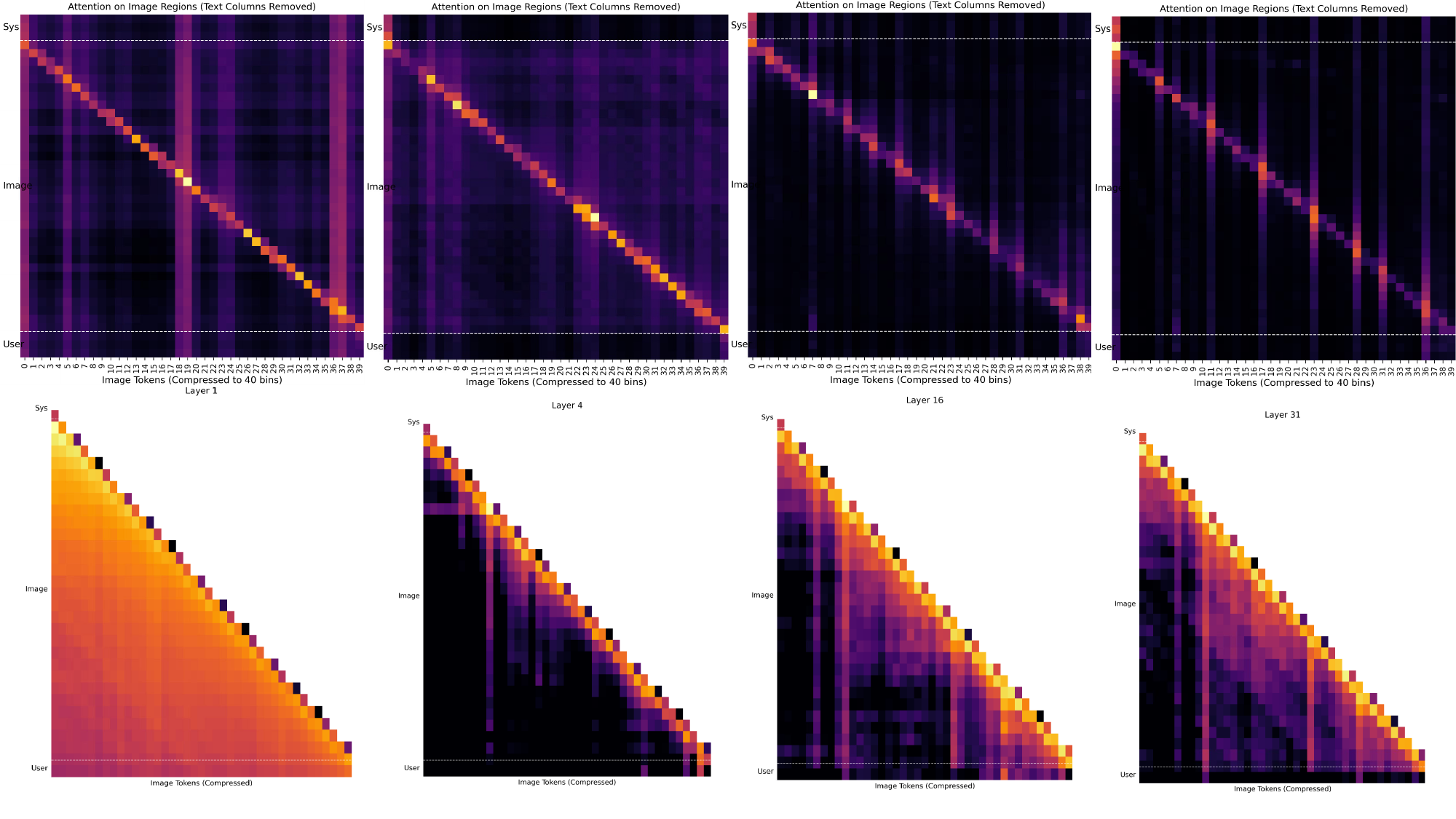}
\caption{\textbf{Comparison of Attention Aggregation Patterns.} \textbf{Top Row (LLaDA-V):} Attention maps remain sparse in early layers (Layers 1--4), with bright vertical bands (semantic aggregation) emerging only after Layer 16. \textbf{Bottom Row (LLaVA\citep{fastv}):} In contrast, the autoregressive baseline exhibits immediate, high-intensity attention aggregation starting from Layer 4, indicating early semantic saturation.}
\label{fig:attn-matrix}
\end{figure}

Specifically, as illustrated in Figure~\ref{fig:attention-vis}, we conduct a quantitative monitoring of attention allocation across different depths. The results reveal that \textbf{Image Tokens} consistently dominate the computational budget, with their attention share rising from \textbf{41.9\%} in Layer 3 to \textbf{53.0\%} in Layer 30. However, this high resource consumption contrasts sharply with their low information density. Taking the attention efficiency of Image Tokens as the baseline (1.0x), the \textbf{Output Tokens}, despite occupying a minimal share of attention (4.7\% at Layer 3 and 10.7\% at Layer 30), exhibit staggering efficiency scores of \textbf{205.7x} and \textbf{366.9x}, respectively. This phenomenon of ``High Allocation, Low Efficiency'' strongly evidences the redundancy within visual representations, indicating that the model wastes substantial computational resources on visual regions that contribute marginally to the semantic reasoning.

Beyond the overall attention allocation ratio, we further visualize the layer-wise text–visual attention matrices at representative depths (Layers 1, 4, 16, and 31) to contrast the behaviors of diffusion-based and autoregressive models. As shown in Figure~\ref{fig:attn-matrix}, the difference is striking:

\textbf{Delayed Aggregation in LLaDA-V (Top Row):} The shallow layers (Layers 1 and 4) exhibit relatively diffuse and dark attention patterns, indicating that the model has not yet formed clear cross-modal aggregation structures. It is only from Layer 16 onward that the attention maps develop pronounced vertical high-intensity bands. This suggests a \textit{delayed semantic aggregation} process, where the model gradually concentrates interactions around a small subset of visually salient tokens as the network deepens.

\textbf{Early Saturation in LLaVA (Bottom Row):} In sharp contrast, the autoregressive LLaVA model displays dense and high-intensity attention patterns as early as Layer 4. The triangular causal masking structure is clearly visible, and visual tokens are heavily attended to immediately. This reflects the "read-and-process" nature of autoregressive decoding, where visual information must be fully integrated upfront to predict the next token.

The divergent behavior observed in LLaDA-V stems fundamentally from its diffusion-style iterative denoising paradigm and bidirectional attention mechanism. Unlike autoregressive (AR) models (e.g., LLaVA, Flamingo) that rely on causal masking and KV cache reuse to accumulate historical context, LLaDA-V recomputes attention over the entire token set at every layer. In AR models, image tokens primarily provide initial grounding; once their information is integrated in early layers, the model shifts focus to language continuity. However, in LLaDA-V, the semantic representations evolve progressively along the depth and do not converge to a compact latent state until the mid-to-late layers. Consequently, the model continues to re-attend to visual tokens during the refinement trajectory, and the formation of stable cross-modal alignment occurs substantially later.

These attention patterns indicate that LLaDA-V does not merely internalize vision features early to detach from them later; rather, visual information acts as an active guide throughout the entire denoising trajectory. The emergence of strong aggregation bands only in the mid-to-late layers (Layer 16+) reflects this vision-guided semantic refinement. This observation provides a strong empirical motivation for our specific pruning strategy: unlike traditional FastV which prunes at Layer 3, we must respect the delayed aggregation behavior of LLaDA-V. Therefore, we aim to eliminate insensitive tokens only after Layer 15, preserving the visually critical ones until semantic convergence is truly achieved.
\section{Method}

We adopt the core idea of FastV to reduce redundant computation in vision--language models by pruning visual tokens during inference. Let the input sequence at layer $l$ be denoted as
\[
X^l = \{x^l_1, x^l_2, \dots, x^l_{N_l}\},
\]
where $N_l$ is the number of tokens at that layer, and $F_l(X^l)$ represents the computational cost (in FLOPs) of processing these tokens. Following FastV, we prune a proportion $R\%$ of visual tokens at a designated layer $K$, producing a reduced sequence $\tilde{X}^K$ of size
\[
|\tilde{X}^K| = (1-R) \cdot N_K,
\]
which yields an estimated TFLOPs reduction
\[
\Delta \text{TFLOPs} \approx \frac{F_K(X^K) - F_K(\tilde{X}^K)}{\text{Total TFLOPs}}.
\]

However, unlike autoregressive models where cross-modal information is progressively aggregated from shallow to deep layers, LLaDA-V employs a bidirectional attention mechanism coupled with a diffusion-style iterative denoising process. Empirical observations from attention heatmaps indicate that salient text--visual aggregation does not occur in the shallow layers; instead, significant alignment emerges only in the middle-to-late layers $M \le l \le L$. Directly pruning early layers as in FastV would risk removing semantically critical tokens before they participate in meaningful cross-modal interactions. To address this, we modify the pruning schedule such that token reduction is applied in the middle-to-late layers of the first denoising step rather than at the earliest layers. Denoting the first diffusion step as $t=1$, the pruned sequence at layer $l$ within this step is
\[
\tilde{X}^l_{t=1} = \text{Prune}(X^l_{t=1}, R), \quad l \in [M, K],
\]
where $\text{Prune}(\cdot, R)$ removes the lowest-contributing $R\%$ visual tokens based on attention scores or FastV saliency metrics. Although these layers are not the absolute shallowest, they already capture sufficient cross-modal alignment to allow safe token reduction. Across subsequent denoising iterations, the computational savings accumulate without substantially impacting the semantic fidelity or the quality of the generated output. Notably, the larger the number of total steps is, the less discrepancy between acceleration effect of different layers in the first step there will be.  In this way, our method aligns the pruning strategy with the delayed attention aggregation in LLaDA-V while retaining the acceleration benefits of FastV.

\section{Experiments}

\subsection{Experimental Setup}

We evaluate our proposed LLaDA-FastV on several standard vision–language benchmarks, including AI2D \citep{kembhavi2016ai2d}, MME \citep{fu2023mme}, MMMU \citep{yue2024mmmu}, MMMU-Pro \citep{yue2024mmmupro}, RealWorldQA \citep{x2024realworldqa}, and ChartQA \citep{masry2022chartqa}. These datasets cover diverse multimodal reasoning tasks, ranging from natural image understanding to complex chart reasoning, allowing us to systematically assess the impact of structured token reduction.

We implement our pruning strategy at different layers ($K \in \{3, 15\}$) and pruning ratios ($P \in \{50\%, 70\% \}$). To quantify efficiency, we report the remaining FLOPs percentage relative to the baseline LLaDA-V model, alongside the reduction rate. Our main results are shown in \textbf{Table~\ref{tab:fastv_llada}}.

\subsection{Main Results Analysis}

Table~\ref{tab:fastv_llada} presents a comprehensive comparison of different pruning configurations. Our analysis yields three key observations:

\textbf{1. Late Pruning Achieves the Best Trade-off.} 
The configuration of $K=15, P=50\%$ emerges as the optimal setting. It reduces the computational cost by approximately \textbf{49\%} while maintaining performance nearly identical to the baseline. Notably, on benchmarks like MMMU (49.89 vs. 48.78) and RealWorldQA (64.84 vs. 63.53), this setting even surpasses the full model, suggesting that removing redundant visual tokens at deeper layers may help reduce noise and improve reasoning focus.

\textbf{2. Sensitivity of Early Layers.} 
Comparing $K=3$ with $K=15$ under the same pruning ratio ($P=50\%$), we observe a significant performance drop in the early-pruning setting. For instance, accuracy on ChartQA plummets from 77.00 ($K=15$) to 35.32 ($K=3$). This indicates that visual tokens in the early layers of LLaDA-V contain critical fine-grained information (e.g., text in charts) that has not yet been fully aggregated. Pruning them too early leads to irreversible information loss.

\textbf{3. Robustness to High Pruning Ratios.} 
Even with a more aggressive pruning ratio of $P=70\%$ at Layer 15, the model retains competitive performance on general VQA tasks (e.g., AI2D: 97.5\% relative performance), while achieving a massive \textbf{63\% reduction} in FLOPs. However, 
aggressive early pruning ($K=3, P=70\%$) results in severe degradation, particularly on detail-oriented tasks like ChartQA and MME.

In summary, our results validate that LLaDA-V's attention mechanism aggregates semantic information in middle-to-late layers. Therefore, a delayed pruning strategy ($K=15$) aligns best with the model's intrinsic behavior, offering substantial efficiency gains with minimal quality loss.

\renewcommand{\per}[1]{\textcolor{black!60}{\scriptsize #1}} 

\begin{table*}[t] 
\centering
\small 
\renewcommand{\arraystretch}{1.2} 
\setlength{\tabcolsep}{0pt} 

\caption{\textbf{Performance comparison of pruning strategies on LLaDA-V.} We compare our proposed \textbf{LLaDA-FastV} (pruning at deeper layers, e.g., K=15) against \textbf{Raw-FastV} (naive application of early pruning, e.g., K=3). The first row shows the raw score, and the second row (in \textcolor{black!60}{gray}) shows the percentage relative to the baseline. The \textbf{Avg.} column represents the average relative performance across all benchmarks. \textbf{FLOPs} indicates the remaining computational cost. Best results among pruned models are \textbf{bolded}.}
\label{tab:fastv_llada}

\begin{tabular*}{\textwidth}{@{\extracolsep{\fill}} l c ccccccc c c }
\toprule
\multirow{2.5}{*}{\textbf{Method}} & \multirow{2.5}{*}{\textbf{Settings}} & \textbf{AI2D} & \textbf{MME} & \textbf{MMMU} & \textbf{\makecell{MMMU\\Pro}} & \textbf{MMStar} & \textbf{\makecell{RealWorld\\QA}} & \textbf{ChartQA} & \textbf{Avg.} & \textbf{FLOPs} \\
\cmidrule(lr){3-9} 
 & & (Acc) & (Score) & (Acc) & (Vis) & (Acc) & (Acc) & (Acc) & (\%) & (Rem./Red.) \\
\midrule

\multicolumn{11}{l}{\textit{\textbf{Baseline}}} \\ 
LLaDA-V & K=-,P=0 
& 77.82 & 2002.5 & 48.78 & 18.38 & 60.32 & 63.53 & 78.16 & \multirow{2}{*}{100.0\%} & 100\% \\
 & 
& \per{100\%} & \per{100\%} & \per{100\%} & \per{100\%} & \per{100\%} & \per{100\%} & \per{100\%} & & \per{-} \\

\midrule

\multicolumn{11}{l}{\textit{\textbf{LLaDA-FastV (Ours)}}} \\

\multirow{2}{*}{} & K=15, P=50 
& \textbf{77.75} & \textbf{1909.0} & \textbf{49.89} & \textbf{17.57} & \textbf{58.67} & \textbf{64.84} & \textbf{77.00} & \multirow{2}{*}{\textbf{98.7\%}} & \textbf{51\%} (49\%$\downarrow$) \\ 
 & 
& \per{99.9\%} & \per{95.3\%} & \per{102.3\%} & \per{95.6\%} & \per{97.3\%} & \per{102.1\%} & \per{98.5\%} & & \\
\addlinespace[4pt]

 & K=15, P=70 
& 75.87 & 1895.0 & 48.22 & 16.47 & 56.80 & 63.79 & 71.76 & \multirow{2}{*}{95.3\%} & 37\% (63\%$\downarrow$) \\
 & 
& \per{97.5\%} & \per{94.6\%} & \per{98.9\%} & \per{89.6\%} & \per{94.2\%} & \per{100.4\%} & \per{91.8\%} & & \\
\addlinespace[4pt]


\midrule

\multicolumn{11}{l}{\textit{\textbf{Raw-FastV}}} \\

\multirow{2}{*}{} & K=3, P=50 
& 71.50 & 1740.0 & 46.78 & 13.64 & 49.05 & 53.73 & 35.32 & \multirow{2}{*}{80.0\%} & 37\% (63\%$\downarrow$) \\
 & 
& \per{91.9\%} & \per{86.9\%} & \per{95.9\%} & \per{74.2\%} & \per{81.3\%} & \per{84.6\%} & \per{45.2\%} & & \\
\addlinespace[4pt]

 & K=3, P=70 
& 66.87 & 1539.0 & 44.44 & 12.14 & 41.74 & 46.80 & 14.48 & \multirow{2}{*}{68.8\%} & 21\% (79\%$\downarrow$) \\
 & 
& \per{85.9\%} & \per{76.9\%} & \per{91.1\%} & \per{66.1\%} & \per{69.2\%} & \per{73.7\%} & \per{18.5\%} & & \\

\bottomrule
\end{tabular*}
\end{table*}

\subsection{Ablation Study: Impact of Pruning Layer and Strategy}

To investigate the optimal configuration for visual token sparsification, we conduct an ablation study comparing our proposed method against two alternative strategies: (1) \textbf{Raw-FastV}, where we apply attention-based pruning at a shallow layer (Layer 3); and (2) \textbf{Random Pruning}, where tokens are discarded randomly to match the computational cost ($P=50\%$).

\begin{figure}[t]
    \centering
    \includegraphics[width=0.85\linewidth]{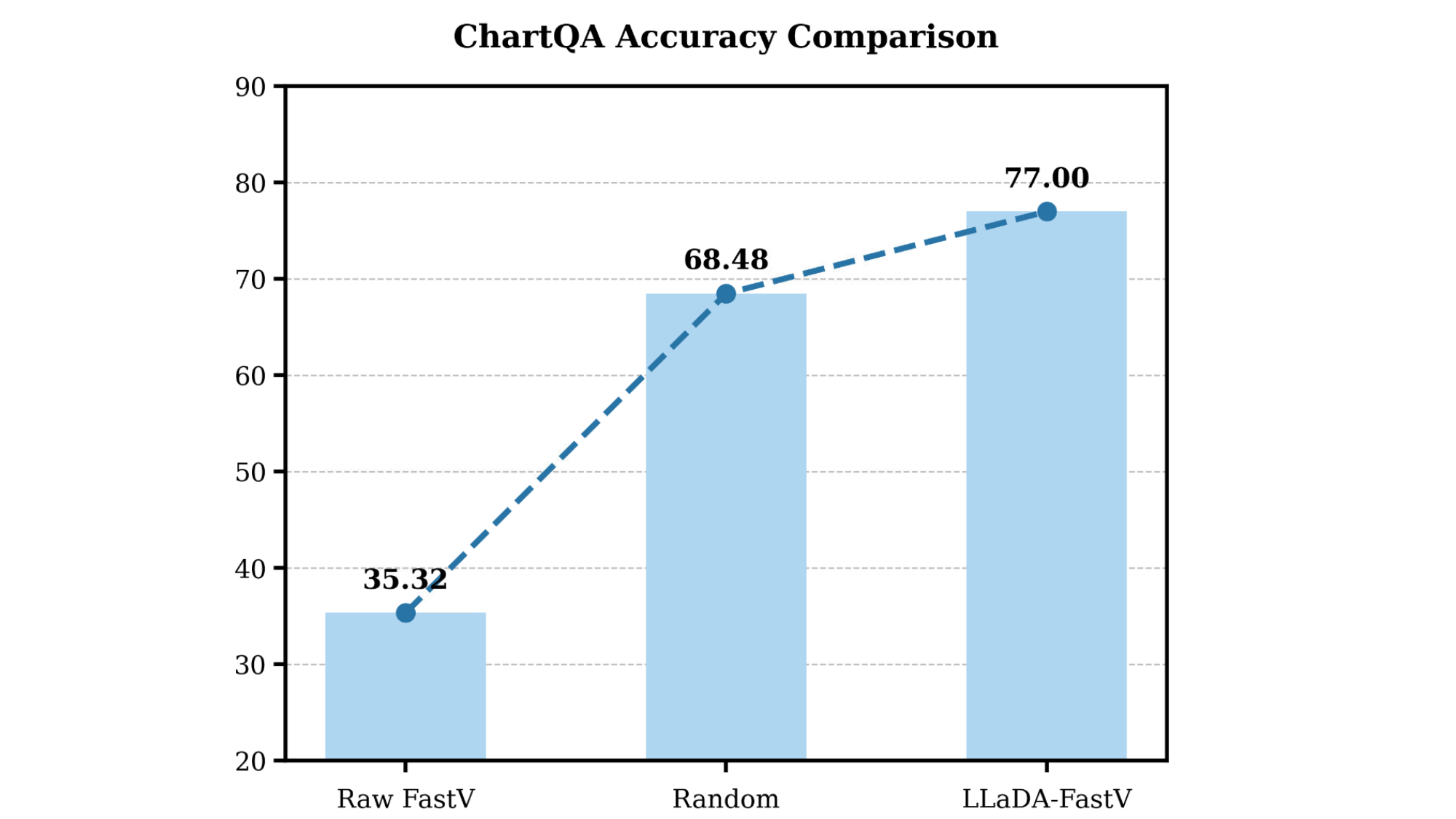} 
    \caption{\textbf{Ablation study on ChartQA.} We compare the impact of pruning strategies (Attention vs. Random) and depth (Layer 3 vs. Layer 15). \textbf{Layer 3 (Raw-FastV)} suffers from immature feature selection, while \textbf{Random} pruning fails to preserve critical details. Our method (\textbf{Layer 15}) effectively balances efficiency and accuracy.}
    \label{fig:ablation}
\end{figure}

The results on the ChartQA benchmark are visualized in \textbf{Figure~\ref{fig:ablation}}. We observe a distinct performance hierarchy that validates our design choices:

\begin{itemize}
    \item \textbf{Failure of Early Pruning (Layer 3):} Applying attention-based pruning at Layer 3 results in a catastrophic performance drop to \textbf{35.32\%}. This suggests that attention maps in shallow layers are still processing low-level features (e.g., texture, edges) and have not yet converged to semantically meaningful regions. Pruning based on these immature attention scores leads to the loss of critical visual information.
    
    \item \textbf{Baseline of Random Pruning:} Random pruning achieves a moderate accuracy of \textbf{68.48\%}. While it outperforms early attention pruning, it still lags significantly behind our method. This indicates that while the model has some redundancy, randomly discarding tokens inevitably removes necessary details required for fine-grained tasks like chart understanding.
    
    \item \textbf{Effectiveness of Deep Pruning (Layer 15):} Our method, applied at Layer 15, achieves the highest accuracy of \textbf{77.00\%}. By deferring sparsification to deeper layers, we leverage the "semantic convergence" of the Vision-Language Model, where the attention mechanism has successfully identified critical tokens. This allows us to aggressively reduce computation while preserving the core reasoning capabilities.
\end{itemize}

\section{Conclusion}

In this work, we conducted an in-depth analysis of the internal attention behavior in LLaDA-V \citep{llada}, revealing that its bidirectional attention mechanism combined with diffusion-style iterative denoising leads to delayed cross-modal information aggregation. Motivated by these findings, we proposed a visual token pruning strategy specifically tailored for LLaDA-V, adopting the core idea of FastV \citep{fastv} while adjusting the pruning schedule: token reduction is applied in the middle-to-late layers of the first denoising step rather than the shallowest layers, thereby preserving critical semantic information while reducing redundant computation. Our method effectively suppresses unnecessary computations without compromising semantic fidelity, providing a practical approach for accelerating large-scale vision–language models. Looking forward, a promising direction is to perform token pruning prior to model input, pruning all visual tokens at once before inference, which has the potential to further amplify acceleration while maintaining high-quality output. Overall, our work demonstrates that visual token pruning is a necessary and effective strategy for efficient inference in LLaDA-V and similar models with complex cross-modal interactions. As a future direction, we suggest exploring pre-inference token pruning across diffusion steps, which may further improve inference efficiency while maintaining semantic integrity.

\bibliography{reference}
\bibliographystyle{unsrt}

\end{document}